\documentclass[10pt,twocolumn,letterpaper]{article}
\pdfoutput=1
\makeatletter
\@namedef{ver@everyshi.sty}{}
\makeatother
\usepackage{iccv}
\usepackage{times}
\usepackage{epsfig}
\usepackage{graphicx}
\usepackage{amsmath}
\usepackage{amssymb}
\usepackage{subcaption}
\usepackage{booktabs}
\usepackage{float}
\usepackage{enumitem}

\usepackage[breaklinks=true,bookmarks=false]{hyperref}

\newcommand{\norm}[1]{\left\lVert#1\right\rVert}
\iccvfinalcopy 

\def\iccvPaperID{6392} 

\newcommand{\mdifgsm}{M-DI$^2$-FGSM}

\newcommand*\rot{\rotatebox{90}}

\usepackage[overlay,absolute]{textpos}
\newcommand\PlaceText[3]{%
\begin{textblock*}{10in}(#1,#2)  
#3
\end{textblock*}
}%

\textblockorigin{-5mm}{0mm}

\begin{document}


\title{Adversarial Examples for Edge Detection:\\They Exist, and They Transfer}

\author{Christian Cosgrove\qquad Alan L. Yuille\\
Department of Computer Science, The Johns Hopkins University\\
Baltimore, MD 21218 USA\\
{\tt\small ccosgro2@jhu.edu\qquad alan.l.yuille@gmail.com}
}

\maketitle

\begin{abstract}
    Convolutional neural networks have recently advanced the state of the art in many tasks including edge and object boundary detection. However, in this paper, we demonstrate that these edge detectors inherit a troubling property of neural networks: they can be fooled by adversarial examples. We show that adding small perturbations to an image causes HED~\cite{SXie:2015:HED:2919332.2920084}, a CNN-based edge detection model, to fail to locate edges, to detect nonexistent edges, and even to hallucinate arbitrary configurations of edges. More surprisingly, we find that these adversarial examples transfer to other CNN-based vision models. In particular, attacks on edge detection result in significant drops in accuracy in models trained to perform unrelated, high-level tasks like image classification and semantic segmentation. Our code will be made public.
\end{abstract}

\section{Introduction}

Edge and contour detection have long played a major role in computer vision. First studied as a low-level function of biological vision~\cite{hubel1962receptive_cat_vision_edge,shapley1973edge_human_vision}, the notion that edge detection can be used to filter out irrelevant lighting and texture information and extract shape information from images dates back to early work in the field~\cite{horn1973binford_line_finder,kittler1983sobel, canny1986computational}. Edge detection has been used as a pre-processing step in many classical vision algorithms~\cite{dollar2013structured, zhu2008hierarchical_feature_learning_uses_edge_detection, Scharstein2002_correspondence, belongie2006matching}. 

The history of edge detection is substantial, and a wide variety of techniques have been developed. Early approaches used hand-crafted features~\cite{kittler1983sobel, canny1986computational}. Later, data-driven methods like~\cite{konishi2003statistical, dollar2013structured} emerged, in which some set of model parameters is automatically tuned on a training dataset in order to reduce false positives. Most recently, convolutional neural networks (CNNs) have been applied to the edge detection problem~\cite{shen2015deepcontour, SXie:2015:HED:2919332.2920084, Bertasius_2015_CVPR_deepedge, liu2016better_edge_detection}. One major success of this line of research is Holistically-Nested Edge Detection (HED), a CNN model that achieves near-human edge detection accuracy on standard datasets~\cite{SXie:2015:HED:2919332.2920084}. This approach has attracted attention for its competitive performance, architectural simplicity, and computational efficiency.

\PlaceText{150mm}{112mm}{$+$}
\PlaceText{130mm}{134mm}{$\downarrow$}
\PlaceText{172mm}{134mm}{$\downarrow$}
\begin{figure}[t]
\captionsetup[subfigure]{labelformat=empty}

\centering

\begin{subfigure}[t]{0.23\textwidth}
\centering
\caption{\textit{``bighorn sheep"}}
\includegraphics[width=\textwidth]{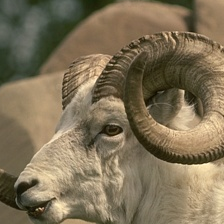}
\label{fig:orig_image_iccv}
\end{subfigure}%
\hfill
\begin{subfigure}[t]{0.23\textwidth}
\centering
\caption{\textit{``Indian elephant"}}
\includegraphics[width=\textwidth]{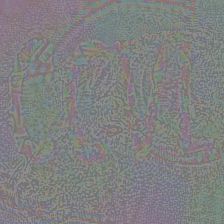}
\label{fig:difference_iccv}
\end{subfigure}

\begin{subfigure}[t]{0.23\textwidth}
\includegraphics[width=\textwidth]{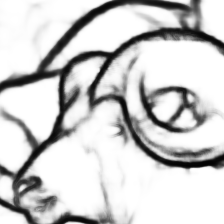}
\label{fig:orig_output_iccv}
\end{subfigure}%
\hfill
\begin{subfigure}[t]{0.23\textwidth}
\includegraphics[width=\textwidth]{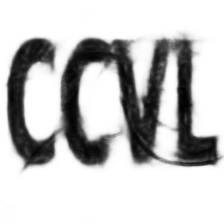}
\label{fig:attacked_output_iccv}
\end{subfigure}

\caption{Adding a small perturbation (right) to an image causes a state-of-the-art edge detection model to produce a contrived pattern. The same perturbation causes a VGG16 model to misclassify the image. We set $y^\text{target}$ to achieve the edge pattern above. Here, $\epsilon = 16$.}
\end{figure}

In recent years, automatic feature learning by CNNs has replaced explicit edge detection for higher-level vision tasks like image classification. However, it is well known that CNNs learn edge-like features implicitly~\cite{krizhevsky2012alexnet}. The Gabor-like filters learned by the earliest layers of CNNs emerge regardless of which dataset or task they are trained on~\cite{yosinski2014transferable_gabor}. In this sense, edge detection is a universal visual task that continues to underlie modern vision systems, albeit implicitly.

Despite CNNs' marked gains in accuracy over classical techniques in domains like classification and semantic segmentation, they are vulnerable to \emph{adversarial examples}. In a variety of tasks~\cite{Szegedy2014intriguing, CXie_2017_ICCV_seg_obj_det}, small perturbations that look like noise to a human can cause the network to produce nonsensical results. In many cases, an attacker can select this perturbation to cause the network to produce any desired output. Worse, some attacks transfer: the same perturbation trained to fool one network sometimes fools similar networks trained on slightly different datasets. 

However, it has not yet been shown whether these adversarial examples are limited to networks trained on ``complex" visual tasks like classification and semantic segmentation, or whether even a CNN trained to perform a low-level task like edge detection is vulnerable. In this paper, we address this question by investigating the degree to which HED suffers from adversarial examples. Adapting existing methods to HED, we find that it is indeed vulnerable to a particular class of adversarial attacks. Altogether, the following results add yet another example to the list of domains where deep neural networks can be fooled.

Just as edge detection is a universal component of many methods in computer vision, we find that adversarial examples for edge detection affect other models, too: \emph{they transfer to higher-level tasks}. In particular, we show that an attack on edges can transfer to models regardless of architecture, training data, and visual task. Without knowing the parameters of a vision model, we can impair that model's accuracy on an image by attacking the edges of the image. The intuition behind these results should be clear: because edge detection is used in CNNs for downstream processing, the CNN will fail to perform higher-level tasks if we can obfuscate these edges.

\section{Related work}

Adversarial examples have primarily been studied in the context of image classification~\cite{Szegedy2014intriguing, Goodfellow2014FGSM, kurakin2017_learning_at_scale}. However, they have also been found to affect networks for object detection~\cite{CXie_2017_ICCV_seg_obj_det}, semantic segmentation~\cite{CXie_2017_ICCV_seg_obj_det, fischer2017adversarial_segmentation}, and natural language processing~\cite{alzantot2018natural_language}. Apart from finding new domains in which adversarial examples exist, much of recent research has focused on devising generic algorithms for generating adversarial examples---\ie, how to synthesize them efficiently and how to improve their success rates. The first work of this kind uses a L-BFGS optimizer to minimize the size of the perturbation subject to the constraint that the network produces the target output~\cite{Szegedy2014intriguing}. The prevalent fast gradient sign method (FGSM)~\cite{Goodfellow2014FGSM} exploits the linearity of the loss function landscape to generate adversarial examples with only first-order information and a single pass of backpropagation. This method has been improved by iterated updates~\cite{kurakin2017_learning_at_scale} and momentum~\cite{Dong_2018_CVPR_momentum}. The literature on defending against these adversarial examples is as rich as the study of the attacks themselves; prominent examples are defensive distillation~\cite{papernot2016distillation}, input transformations~\cite{guo2017input_transformations_defenses}, and adversarial training~\cite{Szegedy2014intriguing, madry2018_adversarial_training}.

The problem of attacking HED is \emph{dense}, meaning that the output space is high-dimensional. In contrast with ImageNet models that have only a 1000-class softmax output, HED produces outputs with tens or hundreds of thousands of edge probabilities---one for each pixel. 
To address the problem of attacking models with high-dimensional outputs,~\cite{CXie_2017_ICCV_seg_obj_det} propose \emph{dense adversary generation} (DAG). During the optimization process, this method ignores pixels (or object proposals) whose output already matches the target. 
DAG was found to be effective in attacking semantic segmentation and object detection models~\cite{CXie_2017_ICCV_seg_obj_det}. We find, however, that generic FGSM attacks suffice to fool HED, so we do not adopt this method. In future work, one could investigate whether DAG leads to more effective attacks on edge detection.

Harmonic Adversarial Attack Method~\cite{Heng2018_harmonic} considers the relationship between edge information and attack quality and transferability. The goal of this work is to maximize the smoothness of the perturbation so that the high-frequency statistics of the image change as little as possible.

Black-box attacks and transferability have been the subject of extensive study since~\cite{Szegedy2014intriguing}. In the black-box setting, the attacker does not have access to the model parameters and architecture; however, the model can be queried to generate an attack. An attack \emph{transfers} if it affects a different model without access to parameters, architecture, or input-output pairs. One approach to generating black-box adversarial examples is to attack a surrogate model trained to mimic outputs from the target model~\cite{papernot2016transferability}. Another is to train a separate network to generate perturbations~\cite{poursaeed2018generative_adversarial_perturbations, baluja2017adversarial_transformation_networks}. Finally, other work studies the transferability of attacks on intermediate layers~\cite{Huang_intermediate_level_adversarial_attack}.

\section{Methods}

    \subsection{Holistically-Nested Edge Detection ~\cite{SXie:2015:HED:2919332.2920084}}

    Like many recent models for semantic segmentation, HED uses a fully-convolutional architecture~\cite{SXie:2015:HED:2919332.2920084}. This means that all of the network's parameters consist of convolution kernels; for this reason, the model is agnostic to input size. HED's convolutional layers are derived from a pretrained VGG16~\cite{simonyan2014vgg16} model and are fine-tuned on the Berkeley Segmentation Data Set (BSDS500)~\cite{amfm_pami2011_bsds500}. A multi-scale architecture and deep supervision are two crucial aspects of the HED method. In particular, HED outputs edge predictions from five different layers of the network, each corresponding to a different scale. During training, each of these \emph{side outputs} is encouraged to match the ground-truth edge map~\cite{SXie:2015:HED:2919332.2920084}.
    
In this paper, we show that despite HED's impressive performance on in-distribution images, this model is easily fooled by adversarial examples. Just like neural network training, the choice of loss function strongly affects the results of an adversarial attack. This is because adversarial attacks are formulated as an optimization problem in the space of images; like learning, generating adversarial examples also uses backpropagation to compute gradients of the loss. Our attack methods optimize a similar cross-entropy loss to that of HED, except for one crucial difference. Consider the loss for side output $m$:
    \begin{equation}
    \begin{split}
        \ell^m(X, y^\text{true}; \theta) = -\frac{1}{2} \sum_{i\,:\,y^\text{true}_i = 1} \log (\hat{y}_i^m)\\-\frac{1}{2} \sum_{i\,:\,y^\text{true}_i = 0} \log (1 - \hat{y}_i^m).
        \label{eq:loss}
    \end{split}
    \end{equation}
Here, $\hat{y}_i^m$ denotes the $i$th pixel of side output $m$, which is a function of $X$ and $\theta$. Unlike HED, we \emph{do not} weigh edges ($y^\text{true}_i = 1$) more strongly than non-edges ($y^\text{true}_i = 0$). Instead, the positive and negative classes are penalized equally. This enables additional types of attacks. In particular, in the class-balanced formulation of HED, using $y^\text{true} = \mathbf{1}$ causes the first term to vanish, since it is proportional to the number of non-edges in the ground truth $y^\text{true}$. This prevents the attack from generating new edges in the image, making so-called \emph{edge activation} attacks impossible. Thus, we use a 1:1 class weighting for all attacks.

Like HED, the overall loss is a linear combination of individual side output losses and a multi-scale fusion term:

\begin{equation}
    \begin{split}
    L(X, y^\text{true}; \theta) = \sum_m \alpha_m \ell^m(X, y^\text{true}; \theta)\\ -\frac{1}{2} \sum_{i\,:\,y^\text{true}_i = 1} \log (\hat{y}_i^\text{fuse})-\frac{1}{2} \sum_{i\,:\,y^\text{true}_i = 0} \log (1 - \hat{y}_i^\text{fuse}),
    \end{split}
    \label{eq:overall_loss}
 \end{equation}
 where $\hat{y}_i^\text{fuse} = \text{sigmoid}(\sum_m h_m \hat{y}^m_i)$. At test time, the final edge prediction is a weighted average of the side outputs and $\hat{y}^\text{fuse}$~\cite{SXie:2015:HED:2919332.2920084}.

    \subsection{Generating adversarial examples}
    In this paper, we apply attacks in the family of fast gradient sign methods (FGSM). These are some of the most studied attack methods~\cite{Goodfellow2014FGSM,kurakin2017_learning_at_scale,kurakin2016physical_world,Dong_2018_CVPR_momentum}, and they require relatively little computation when compared with methods like L-BFGS~\cite{Goodfellow2014FGSM}. In the following section, we describe a few relevant examples of fast gradient sign methods, adopting the notation of~\cite{CXie_2018_diversity}.
    
    The original FGSM~\cite{Goodfellow2014FGSM} generates an adversarial perturbation using the gradient of the loss \begin{equation}
        X^\text{adv} = X + \epsilon\,\text{sign}(\nabla_X L(X, y^\text{true}; \theta)),
    \end{equation}
    where $y^\text{true}$ is the ground-truth edge map. FGSM can be extended to the iterative fast gradient sign method (I-FGSM)~\cite{kurakin2017_learning_at_scale} and the momentum iterative fast gradient sign method (MI-FGSM)~\cite{Dong_2018_CVPR_momentum}, the latter of which uses the update rule
    \begin{align}
        g_{n+1} &= \mu g_n + \frac{\nabla_X L(X^\text{adv}_n, y^\text{true}; \theta)}{\norm{\nabla_X L(X^\text{adv}_n, y^\text{true}; \theta)}_1} \label{eq:mi_fgsm1}\\
        X^\text{adv}_{n+1} &= \text{Clip}_X^\epsilon \left[X^\text{adv}_n + \alpha\,\text{sign}(g_{n+1})\right],\label{eq:mi_fgsm2}
    \end{align}
    where $\epsilon\geq\norm{X - X^\text{adv}}_\infty$ measures the size of the perturbation and the momentum $\mu$ and step size $\alpha$ are attack parameters. In this paper, all attacks are based on MI-FGSM.
    
    In transferability studies, \cite{CXie_2018_diversity} showed that introducing \emph{input diversity transformations} makes attack perturbations more likely to transfer across architectures. Like data augmentation, the input image $X$ is randomly resized during the optimization process. Following this approach, we test \mdifgsm, a modified version of MI-FGSM, in our transfer experiments. This replaces the update in Eq.~\ref{eq:mi_fgsm1} with
    \begin{align}
        g_{n+1} &= \mu g_n + \frac{\nabla_X L(T(X^\text{adv}_n), y^\text{true}; \theta)}{\norm{\nabla_X L(T(X^\text{adv}_n), y^\text{true}; \theta)}_1}
        \label{eq:mi_fgsm_input_diversity}
    \end{align}
    where
    \begin{align}
        T(X) = \begin{cases}
        \text{resize}(X) & \text{with probability } 1/2 \\
        X & \text{otherwise}\\
        \end{cases}
        \label{eq:resize_function}
    \end{align}
    The transformation function $\text{resize}(X)$ first down-scales the image to a rectangle with random dimensions $(w,h)$---where $w,h \sim \text{Uniform}(0, 300)$---then randomly pads the boundaries of the image with black pixels to restore it to its original size.
    
    After perturbing the image, it is possible that pixel intensities of $X^\text{adv}$ leave the valid range [0, 255]. To deal with this, we simply clip pixel intensities to [0, 255] after adding the perturbation. Although this can destroy some of the perturbation, \cite{CXie_2017_ICCV_seg_obj_det} find that the effect is negligible for small $\epsilon$, so we adopt this practice.

    \begin{figure*}
    
\captionsetup[subfigure]{justification=centering}
    \centering
    \begin{subfigure}[t]{0.19\textwidth}
    \centering
    \includegraphics[width=\textwidth]{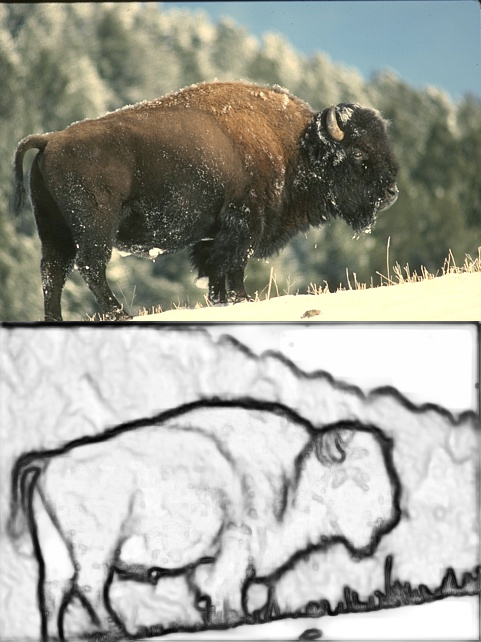}
    \caption{no attack}
    \label{fig:unattacked_output}
    \end{subfigure}
    \hfill
    \begin{subfigure}[t]{0.19\textwidth}
    \centering
    \includegraphics[width=\textwidth]{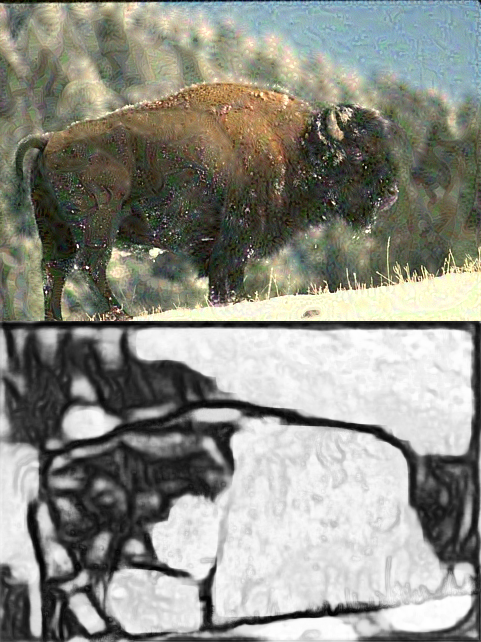}
    \caption{MI-FGSM}
    \label{fig:untargeted_attack}
    \end{subfigure}
    \hfill
    \begin{subfigure}[t]{0.19\textwidth}
    \centering
    \includegraphics[width=\textwidth]{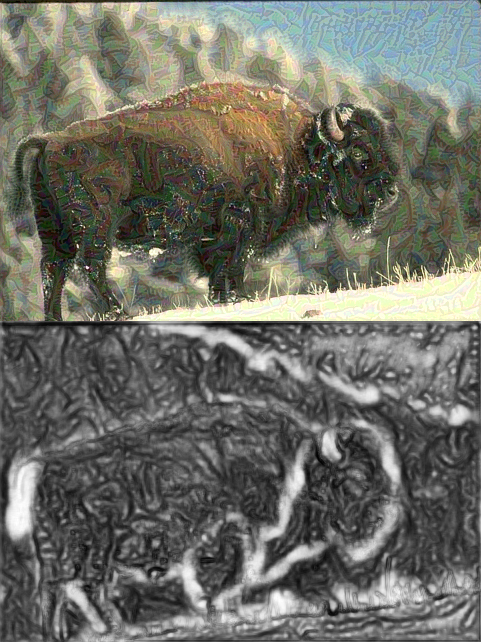}
    \caption{I-MI-FGSM\\$y^\text{target} = \mathbf{1} - y^\text{true}$}
    \label{fig:inversion_attack}
    \end{subfigure}
    \hfill
    \begin{subfigure}[t]{0.19\textwidth}
    \centering
    \includegraphics[width=\textwidth]{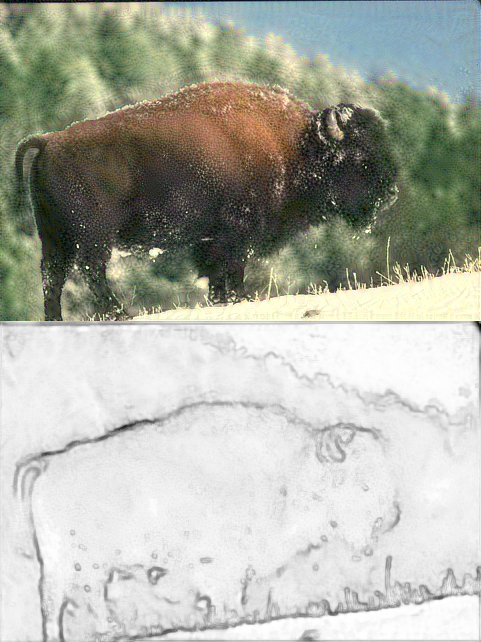}
    \caption{S-MI-FGSM\\$y^\text{target} = \mathbf{0}$}
    \label{fig:suppression_attack}
    \end{subfigure}
    \hfill
    \begin{subfigure}[t]{0.19\textwidth}
    \centering
    \includegraphics[width=\textwidth]{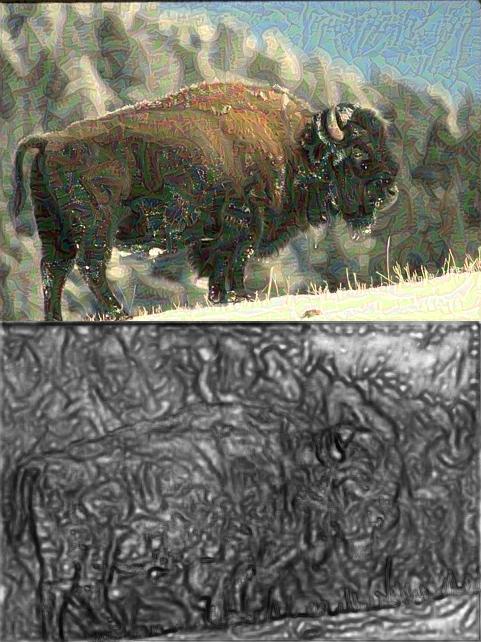}
    \caption{A-MI-FGSM\\$y^\text{target} = \mathbf{1}$}
    \label{fig:activation_attack}
    \end{subfigure}

    \caption{Figure~\protect{\ref{fig:unattacked_output}} is an unaltered image from the BSDS500 test set and the output of HED. Attack~\protect{\ref{fig:untargeted_attack}} uses the untargeted MI-FGSM optimizer (Eqs.~\protect{\ref{eq:mi_fgsm1}} and~\protect{\ref{eq:mi_fgsm2}}). Inverse-target (Figure~\protect{\ref{fig:inversion_attack}}), suppression (Figure~\protect{\ref{fig:suppression_attack}}), and activation attacks (Figure~\protect{\ref{fig:activation_attack}}) use the targeted MI-FGSM optimizer (Eqs.~\protect{\ref{eq:mi_fgsm1_targeted}} and~\protect{\ref{eq:mi_fgsm2_targeted}}). Here, we use $\epsilon=16$ and 10 iterations.}
    \label{fig:attack_variants}
    \end{figure*}
    
    \subsection{Targeted attacks}
    Up to this point, we have only discussed so-called \emph{untargeted} attacks, which maximize the original training loss. Adversarial attacks also come in a \emph{targeted} form that \emph{minimizes}, rather than maximizes, a \emph{modified} loss. For example, targeted MI-FGSM has the update rule
    \begin{align}
        g_{n+1} &= \mu g_n + \frac{\nabla_X L(X^\text{adv}_n, y^\text{target}; \theta)}{\norm{\nabla_X L(X^\text{adv}_n, y^\text{target}; \theta)}_1} \label{eq:mi_fgsm1_targeted}\\
        X^\text{adv}_{n+1} &= \text{Clip}_X^\epsilon \left[X^\text{adv}_n - \alpha\,\text{sign}(g_{n+1})\right], \label{eq:mi_fgsm2_targeted}
    \end{align}
    where $y^\text{target}$ is the desired output of the network. Note that $y^\text{true}$ has been changed to $y^\text{target}$ and the sign in front of $\alpha$ is now negative. This leads to four main attack variants, the last three of which are targeted:
    
    \begin{enumerate}
        \item[$\mathcal{U}$] \textbf{Untargeted attack}. Use Eqs.~\ref{eq:mi_fgsm1} and~\ref{eq:mi_fgsm2}.
        \item[$\mathcal{S}$] \textbf{Suppression attack}. The objective is to lower the probability of edges throughout the image. This corresponds to setting $y^\text{target} = \mathbf{0}$.
        \item[$\mathcal{A}$] \textbf{Activation attack}. The objective is to increase the probability of edges throughout the image. This corresponds to setting $y^\text{target} = \mathbf{1}$.
        \item[$\mathcal{I}$] \textbf{Inverse-target attack}. The objective is to minimize the loss on the \emph{inverted} ground truth label, using $y^\text{target} = \mathbf{1} - y^\text{true}$. This is an alternative to untargeted attacks.
    \end{enumerate}

    \begin{table}[]
    \centering
    \begin{tabular}{@{}rlllll@{}}
    \toprule
    \rot{variant}&\rot{attack name}      & \rot{targeted?} & \rot{optimizer} & \rot{$y^\text{target}$} & \rot{input diversity?} \\ \midrule
    $\mathcal{U}$ &MI-FGSM    &           & Eqs.~\ref{eq:mi_fgsm1}, \ref{eq:mi_fgsm2} & n/a          &                                                \\
    $\mathcal{S}$ &\textbf{S}-MI-FGSM  & \checkmark& Eqs.~\ref{eq:mi_fgsm1_targeted}, \ref{eq:mi_fgsm2_targeted}     & $\mathbf{0}$          &                                 \\
    $\mathcal{A}$ &\textbf{A}-MI-FGSM  & \checkmark& Eqs.~\ref{eq:mi_fgsm1_targeted}, \ref{eq:mi_fgsm2_targeted}     & $\mathbf{1}$          &                                 \\
    $\mathcal{I}$ &\textbf{I}-MI-FGSM  & \checkmark& Eqs.~\ref{eq:mi_fgsm1_targeted}, \ref{eq:mi_fgsm2_targeted}     & $\mathbf{1} - y^\text{true}$ &                                                \\
    $\mathcal{U}$ &\mdifgsm   &           & Eqs.~\ref{eq:mi_fgsm_input_diversity}, \ref{eq:mi_fgsm2_targeted}     & n/a   & \checkmark                           \\
    $\mathcal{S}$ &\textbf{S}-\mdifgsm & \checkmark& Eqs.~\ref{eq:mi_fgsm_input_diversity}, \ref{eq:mi_fgsm2_targeted}     & $\mathbf{0}$          &   \checkmark                             \\
    $\mathcal{A}$ &\textbf{A}-\mdifgsm & \checkmark& Eqs.~\ref{eq:mi_fgsm_input_diversity}, \ref{eq:mi_fgsm2_targeted}     & $\mathbf{1}$          & \checkmark                               \\
    $\mathcal{I}$ &\textbf{I}-\mdifgsm & \checkmark& Eqs.~\ref{eq:mi_fgsm_input_diversity}, \ref{eq:mi_fgsm2_targeted}     & $\mathbf{1} - y^\text{true}$ & \checkmark                           \\\bottomrule
    \end{tabular}
    \caption{Attack variants. \textbf{No prefix} ($\mathcal{U}$): \emph{untargeted} attacks. \textbf{S- prefix} ($\mathcal{S}$): \emph{suppression} attacks. \textbf{A- prefix} ($\mathcal{A}$): \emph{activation} attacks. \textbf{I- prefix} ($\mathcal{I}$): \emph{inverse-target} attacks. The last four attacks are the same as the first four, except they use input diversity transformations (Eqs.~\protect{\ref{eq:mi_fgsm_input_diversity}},~\protect{\ref{eq:resize_function}}).}
    \label{table:attack_variant_clarifications}
    \end{table}
    
    \subsection{Evaluation}
    We evaluate our approach on the test set of the BSDS500 dataset~\cite{amfm_pami2011_bsds500}. This consists of 200 images with ground-truth boundary annotations. Like~\cite{SXie:2015:HED:2919332.2920084}, evaluation is performed using the fixed-contour threshold F-score (ODS).
    We perform the same standard non-maximal-suppression procedure as~\cite{canny1986computational, SXie:2015:HED:2919332.2920084} before evaluating outputs. To measure the effectiveness of the attack, we compare the mean ODS
    of HED outputs on unattacked images and outputs on attacked images. 
    
    To compensate for not using a class-balanced loss (Eq.~\ref{eq:loss}), we apply a morphological thickening operation (radius of 3 pixels) to the ground-truth labels $y^\text{true}$ for untargeted attacks.
\section{Edge attack experiments}
\label{section:edge_attacks}
In the following experiments, we evaluate these attack variants on BSDS500. All attacks are run for 10 iterations with $\epsilon = 16$. We fix $\mu = 0.5$ and $\alpha = 2$. In Table~\ref{table:f_vs_epsilon}, we see that all methods decrease the ODS F-score of HED, with I-MI-FGSM being the most effective attack. For other methods, we find that the combination of side-output averaging and non-maximal suppression protects HED against major drops in accuracy, even though the raw output of the network changes (see Figure~\ref{fig:attack_variants}).

It is worth noting that these attacks are less successful at suppressing edges than activating non-edges. In Figure~\ref{fig:suppression_attack}, notice that the boundary of the buffalo is still detected, albeit with much lower probability. This may be due to the weighting of the loss function in Eq.~\ref{eq:loss}, as edges are less frequent than non-edges so they contribute less to the loss function. In general, we find that the attacks typically fail to suppress unambiguous edges (\eg, the high-contrast leg in Figure~\ref{fig:attack_variants}) and fool those that require more global context to detect (\eg, the boundary between the mountain and the sky in Figure~\ref{fig:attack_variants}). 

\begin{figure}[t]
\centering
   \includegraphics[width=\linewidth]{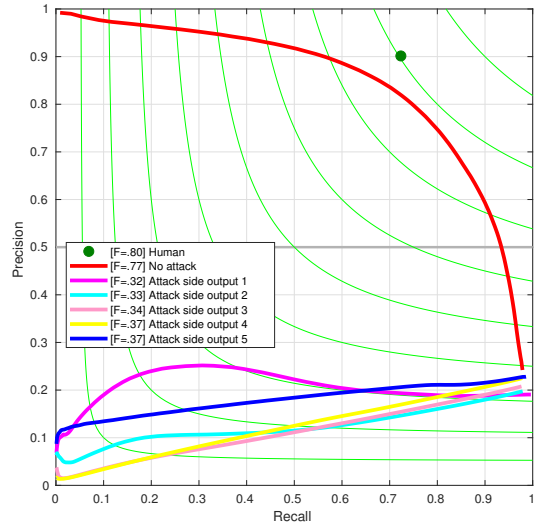}
   \caption{Precision-recall curves from I-MI-FGSM attacks on different side outputs.
   The ``inverted" precision-recall curves are due to the fact that I-MI-FGSM makes the network more likely to classify edges as non-edges and non-edges as edges.
   }
\label{fig:side_output_precision_recall}
\end{figure}

\begin{table}[t]
\centering
\begin{tabular}{@{}rllll@{}}
\toprule
$\epsilon$ & MI-FGSM & I-MI-FGSM & S-MI-FGSM & A-MI-FGSM \\ \midrule
0       & 0.775   & 0.775     & 0.775     & 0.775     \\
1       & \textbf{0.720}   & 0.728     & 0.752     & 0.756     \\
2       & 0.680   & \textbf{0.637}     & 0.731     & 0.726     \\
4       & 0.636   & \textbf{0.445}     & 0.702     & 0.684     \\
8       & 0.588   & \textbf{0.332}     & 0.645     & 0.642     \\
16      & 0.545   & \textbf{0.312}     & 0.573     & 0.580     \\ \bottomrule
\end{tabular}
\caption{BSDS500 test ODS F-score as a function of attack magnitude $\epsilon$ for each attack variant.}
\label{table:f_vs_epsilon}
\end{table}
We also investigate the effect of attacking HED's individual side outputs. Instead of optimizing the loss in Eq.~\ref{eq:overall_loss}, we simply optimize the single-side-output loss in Eq.~\ref{eq:loss} for $m = 1,\ldots,5$. As shown in Figure~\ref{fig:side_output_precision_recall}, attacking side outputs 2-5 is more effective than attacking side output 1. Two possible explanations for this are (1) when attacking deeper layers, more of the network's parameters are available for the attack to exploit, and (2) later side outputs have larger receptive fields, so each edge output value is a function of more degrees of freedom in the input. These findings agree with the previous paragraph and show that the deeper, non-local layers of HED are the most vulnerable to attack.

\begin{table}[t]
\centering
\begin{tabular}{@{}rllll@{}}
\toprule
           & $\ell^2$    & SSIM  & ESSIM & Laplacian \\ \midrule
Unattacked & 0.000 & 1.000 & 1.000     & 37.234     \\
MI-FGSM    & 0.090 & 0.760 & 0.342     & 45.441     \\
I-MI-FGSM    & 0.107 & 0.684 & 0.281     & 47.875    \\
S-MI-FGSM  & 0.096 & 0.754 & 0.318     & 50.589     \\
A-MI-FGSM  & 0.106 & 0.675 & 0.297     & 48.708     \\ \bottomrule
\end{tabular}
\caption{Measurements of image degradation due to various attacks. Each metric is computed on every image in the BSDS500 test set; mean values are reported. $\ell^2$ corresponds to the normalized $\ell^2$-norm, \ie $\norm{X - X^\text{adv}}_2/\norm{X}_2$. \emph{SSIM}~\protect{\cite{wang2004ssim}} is a prevalent measurement of image degradation. \emph{ESSIM}~\protect{\cite{Heng2018_harmonic}}, or Edge-SSIM, is obtained by applying SSIM to the Laplacian maps of each image. \emph{Laplacian} corresponds the mean absolute value of the Laplace operator evaluated on the attacked image.}
\label{table:degradation_metrics}
\end{table}


Some of the perturbations generated by our method can be perceived by the eye; we quantify this perceptibility using image quality metrics. In Table~\ref{table:degradation_metrics}, we see that it is possible to differentiate attacked images from unattacked images using metrics for image quality degradation. This makes sense in the context of earlier findings that FGSM results in loss of SSIM and Edge-SSIM scores~\cite{Heng2018_harmonic}. Note that the mean absolute value of the Laplacian is higher for attacked images, but this effect is small. Altogether, these visual and statistical discrepancies suggest that it may be possible to detect if the edges of an image have been attacked. We leave this as an open research question.

\section{Transferability}
The following section is concerned with how attacks on HED transfer to higher-level vision tasks. These results are more surprising than those of Section~\ref{section:edge_attacks}, since they show that edge-based attacks have implications beyond edge detection. We consider two important tasks: image classification (a high-level task) and semantic segmentation (a low- and high-level task). It is worth investigating both since edges are a low-level feature, and altering edges might have different effects on networks trained to perform high-level tasks compared to those trained to perform low-level tasks. (Edge detection and semantic segmentation differ in that edges may not correspond to object boundaries; the related task of \emph{object boundary detection}~\cite{premachandran2015pascal_boundaries} corresponds more closely to semantic segmentation.)
\subsection{Classification}
First, we study the effect of edge-based attacks on ImageNet classification~\cite{deng2009imagenet}. For each model, we report the top-1 classification accuracy on the validation set \emph{before attack} and \emph{after attack}. All classifiers use a standard implementation and publicly-available pretrained weights. We test VGG16~\cite{simonyan2014vgg16}, the architecture that HED is based on. To investigate cross-architecture transfer, we also consider models from the ResNet family~\cite{he2016resnet}. We use the same momentum and step size parameters from the previous experiments. When not stated, we set $\epsilon=16$, which yields a slight perturbation. Because we do not have ground-truth edge annotations for the ImageNet dataset, we use the output of HED for $y^\text{true}$. This is necessary for untargeted attacks like MI-FGSM (Eqs.~\ref{eq:mi_fgsm1} and~\ref{eq:mi_fgsm2}).


\begin{table*}[hbt!]
\begin{center}
\begin{tabular}{@{}r|lllllllll@{}}
\toprule
          & none & \shortstack{MI-\\FGSM} & \shortstack{M-DI$^2$-\\FGSM} & \shortstack{I-MI-\\FGSM} & \shortstack{I-M-DI$^2$-\\FGSM} & \shortstack{S-MI-\\FGSM} & \shortstack{S-M-DI$^2$-\\FGSM} & \shortstack{A-MI-\\FGSM} & \shortstack{A-M-DI$^2$-\\FGSM}    \\ \midrule
VGG16     & 71.264     & 25.184  & 25.698  & 15.330 & 14.830    & 26.554    & 26.042       & 15.856    & \textbf{14.332} \\
ResNet18  & 68.932     & 36.748  & 36.944  & 29.380 & 28.902   & 36.480     & 35.140        & 30.310     & \textbf{28.662} \\
ResNet34  & 72.766     & 43.910  & 44.696  & 35.630 & \textbf{34.202}  & 43.448    & 42.720        & 36.374    & 34.544 \\
ResNet50  & 75.586     & 46.202  & 46.818  & 36.772 & 35.574  & 45.518    & 45.554       & 37.152    & \textbf{35.002} \\
ResNet101 & 77.122     & 50.354  & 51.128  & 41.132 & \textbf{40.094}  & 49.704    & 49.206       & 42.346    & 40.644 \\
ResNet152 & 78.018     & 53.418  & 54.030  & 45.106 & \textbf{43.722}   & 53.146    & 52.816       & 45.870     & 43.734 \\ \bottomrule
\end{tabular}
\end{center}
\caption{The top-1 ImageNet accuracy of classification models under various edge-based attacks.}
\label{table:transferability_classification}
\end{table*}

\begin{table*}[hbt!]
\centering
\begin{tabular}{@{}rllllll@{}}
\toprule
Side output: & VGG16           & ResNet18        & ResNet34        & ResNet50        & ResNet101       & ResNet152       \\ \midrule
1                   & 39.042          & 46.376          & 51.456          & 52.33           & 55.262          & 58.832          \\
2                   & 25.67           & 40.244          & 48.122          & 47.146          & 51.97           & 54.83           \\
3                   & \textbf{17.352} & \textbf{32.87}  & 40.584          & \textbf{38.896} & \textbf{42.946} & \textbf{46.758} \\
4                   & 20.884          & 34.642          & \textbf{40.38}  & 40.94           & 46.288          & 49.984          \\
5                   & 22.896          & 36.424          & 41.19           & 44.03           & 48.146          & 51.996          \\
All                 & \textbf{14.332} & \textbf{28.662} & \textbf{34.544} & \textbf{35.002} & \textbf{40.644} & \textbf{43.734} \\ \bottomrule
\end{tabular}
\caption{Classification transferability results when A-\mdifgsm\;is applied to one of the side outputs of HED. Of all of the individual side outputs, the third one is the best to attack (with the exception of ResNet34). However, attacking all side outputs simultaneously (by optimizing Eq.~\protect{\ref{eq:overall_loss}} directly) still transfers the best.}
\label{table:side_output_transferability}
\end{table*}

\begin{table*}[hbt!]
\begin{center}
\begin{tabular}{@{}r|lllll@{}}
\toprule
          & unattacked & A-MI-FGSM       & \begin{tabular}[c]{@{}l@{}}A-MI-FGSM\\ (permuted)\end{tabular} & \begin{tabular}[c]{@{}l@{}}VGG16\\ MI-FGSM\end{tabular} & \begin{tabular}[c]{@{}l@{}}ResNet34\\ MI-FGSM\end{tabular} \\ \midrule
VGG16     & 71.264     & 15.856          & 65.112                                                         & \textbf{2.274}                                          & 65.774                                                     \\
ResNet18  & 68.932     & \textbf{30.31}  & 64.262                                                         & 63.97                                                   & 59.784                                                     \\
ResNet34  & 72.766     & 36.374          & 68.982                                                         & 68.546                                                  & \textbf{0.536}                                             \\
ResNet50  & 75.586     & \textbf{37.152} & 71.424                                                         & 71.378                                                  & 67.87                                                      \\
ResNet101 & 77.122     & \textbf{42.346} & 73.718                                                         & 73.7                                                    & 70.34                                                      \\
ResNet152 & 78.018     & \textbf{45.87}  & 74.878                                                         & 74.946                                                  & 72.046                                                     \\ \bottomrule
\end{tabular}
\end{center}
\caption{A comparison of top-1 ImageNet classification accuracies on images attacked with A-MI-FGSM and white-box MI-FGSM. The third column is obtained by directly attacking VGG16 using MI-FGSM, then evaluating all six models on the perturbed images. As shown, white-box attacks on VGG16 and ResNet34 are more effective than edge-based A-MI-FGSM, but they do not transfer as well to the other models, unlike edge-based attacks. The third column, which highlights the importance of the structure of the perturbation, is obtained by permuting the pixels returned by A-MI-FGSM, in a similar manner to~\protect{\cite{CXie_2017_ICCV_seg_obj_det}}.}
\label{table:transferability_white_box_comparison}
\end{table*}

As shown in Table~\ref{table:transferability_classification}, edge-based attacks do transfer to ImageNet classifiers. The most significant decline is in VGG16, whose accuracy drops from 71.264\% to 14.332\% on images attacked with A-\mdifgsm. This is not much of a surprise: VGG16 and HED share the same architecture, and HED is pretrained with VGG16 weights, so one might expect attacks on HED to transfer to VGG16. More unexpectedly, however, the same perturbations also cause the accuracy of ResNet models to drop precipitously. The effect is greater the shallower the model---ResNet18 suffers a 40-point drop from A-\mdifgsm\;whereas ResNet152 only has a 34-point drop---but in all cases the reduction is consequential. This drop does not occur when the pixels of the perturbation are randomly permuted (column 3 of Table~\ref{table:transferability_white_box_comparison}), showing that the structure of the perturbation matters.

Again in Table~\ref{table:transferability_classification}, observe that the edge activation attack A-\mdifgsm\; and the inverse-target attack I-\mdifgsm\;transfer the best to classification networks (columns 5, 9). In part, this is due to the small boost in transferability that comes from input diversity transformations. Nonetheless, the effect is small, and the same techniques without input diversity transformations transfer almost as well (\eg, compare columns 8 and 9). Overall, it appears that edge activation attacks transfer to classification much better than other types of edge-based attacks.

To give a better understanding of transferability of attacks on edge detection, Table~\ref{table:transferability_white_box_comparison} compares the drop in classification accuracy due to edge-based attacks versus white-box attacks on the same models. In particular, we attack VGG16 and ResNet34 directly, using the same attack method (MI-FGSM) and identical parameters ($\epsilon=16$, $\mu=0.5$, $\alpha=2$, 10 iterations). However, instead of using gradients from HED, we use gradients of the cross-entropy loss from the ImageNet ground truth like standard white-box classification attacks~\cite{Goodfellow2014FGSM}. The white-box MI-FGSM attacks on VGG16 and ResNet34 are highly effective on their respective models (both lead to less than 3\% accuracy). However, unlike edge attacks, the perturbations from these attacks do not transfer to the other models.

We conduct an additional experiment to see if white-box adversarial examples for classification transfer to edge detection. Using a ResNet18 model, we generate adversarial examples for each image in the BSDS500 test set using MI-FGSM. Here, we set $y^\text{true}$ to the output of ResNet18 on the BSDS500 images. We find no difference in HED F-scores on the perturbed images versus unperturbed images, indicating that the transferability only works in one direction.

Table~\ref{table:side_output_transferability} shows the transferability of attacks on different HED side outputs. Here, we choose A-\mdifgsm, which we found has the highest transferability (Table~\ref{table:transferability_classification}). We observe that attacking side output 3 transfers the best to classification (\ie, optimizing the loss in Eq.~\ref{eq:loss} for $m=3$). This result mimics findings that attacking intermediate layers of classifiers---rather than output layers---leads to greater transferability of adversarial examples~\cite{Huang_intermediate_level_adversarial_attack}. However, attacking the multi-scale loss of Eq.~\ref{eq:overall_loss} still transfers better than attacking any individual side output. 

\subsection{Texture bias and transferability}
\label{section:stylized_imagenet}
To address the question of why edge-based attacks transfer to ImageNet classification, we evaluate these attacks on models that have been explicitly trained to ignore texture. It has been shown that ImageNet training induces texture bias in CNNs~\cite{geirhos2018stylized_imagenet}. The Stylized-ImageNet dataset~\cite{geirhos2018stylized_imagenet} consists of ImageNet training examples that have transformed using AdaIN style transfer~\cite{Huang_2017_adain_style_transfer}. This style transfer removes low-level texture information but preserves global shape. The authors of this dataset show that merely training on Stylized-ImageNet rather than ImageNet reduces a CNN's reliance on texture to classify images~\cite{geirhos2018stylized_imagenet}. In this experiment, we investigate whether this reduced reliance on texture improves robustness to edge-based attacks.

To answer this question, we compare the ImageNet validation accuracy of a ResNet50 model trained on ImageNet and a ResNet50 model trained jointly on ImageNet and Stylized-ImageNet under various edge-based attacks. According to the first row of Table~\ref{table:stylized_imagenet}, both models have roughly the same performance on unattacked images (75.586\% versus 74.074\%). However, on edge-based adversarial examples, the two models' accuracy differs considerably.

\begin{table}[hbtp!]
\begin{center}
\begin{tabular}{@{}r|ll@{}}
\toprule
            & \begin{tabular}[c]{@{}l@{}}ImageNet\\ (texture-biased)\end{tabular} & \begin{tabular}[c]{@{}l@{}}ImageNet\\   + Stylized-ImageNet\\ (shape-biased)\end{tabular} \\ \midrule
unattacked  & \textbf{75.586}   & 74.074\\
MI-FGSM     & \textbf{46.202}   & 40.952\\
S-MI-FGSM     & 45.518   & \textbf{51.778}\\
A-M-DI$^2$-FGSM & 35.002 & \textbf{48.786}\\ \bottomrule
\end{tabular}
\end{center}
\caption{Top-1 ImageNet validation accuracies of a ResNet50 model trained on ImageNet and both the ImageNet and the Stylized-ImageNet datasets. Decreasing texture bias by training on Stylized-ImageNet improves robustness to suppression and activation attacks like S-MI-FGSM and A-\mdifgsm, but it reduces robustness to MI-FGSM.}
\label{table:stylized_imagenet}

\end{table}
In Table~\ref{table:stylized_imagenet}, note that training on stylized images improves robustness to edge suppression and activation attacks, but it decreases robustness to untargeted MI-FGSM edge attacks. In particular, the Stylized-ImageNet model achieves a 6-point lower accuracy on MI-FGSM adversarial examples than the ImageNet-trained model. Since this model is biased towards shape, this suggests that MI-FGSM targets shape information more than S-MI-FGSM or A-\mdifgsm. On the other hand, on adversarial examples generated with S-MI-FGSM, the shape-biased model achieves a 6-point higher accuracy, and on those generated with A-\mdifgsm, it improves by 13 points. This suggests that the edge suppression and edge activation attacks obfuscate texture more than untargeted edge attacks.

\subsection{Semantic segmentation}

In addition to classification, we also study whether adversarial examples for edge detection transfer to semantic segmentation. Like those for classification, adversarial examples for semantic segmentation have also been shown to transfer between deep network architectures~\cite{CXie_2017_ICCV_seg_obj_det}.
DeepLabv3+~\cite{deeplabv3plus2018} is a state-of-the-art CNN model for semantic segmentation. We test a model provided by the authors of the paper~\cite{deeplabv3plus2018} that is pretrained on MS-COCO~\cite{lin2014coco} and on augmented training examples from PASCAL VOC 2012 ~\cite{pascal-voc-2012}. To evaluate the transferability of edge-based attacks, we compare the mean intersection over union (mIOU) of DeepLabv3+ on unperturbed and perturbed images from the PASCAL VOC 2012 validation set. 

\begin{figure}[]
\captionsetup[subfigure]{labelformat=empty}
\centering

\includegraphics[width=2mm]{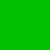} sofa \quad 
\includegraphics[width=2mm]{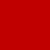} chair \quad 
\includegraphics[width=2mm]{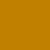} table \quad 
\includegraphics[width=2mm]{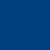} tv
\vspace{2mm}

\begin{subfigure}[t]{0.23\textwidth}
\centering
\includegraphics[width=\textwidth]{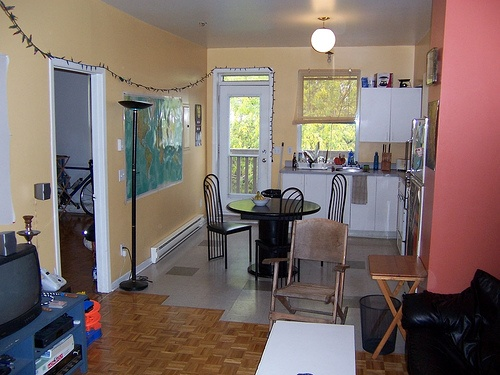}
\label{fig:unattacked_deeplab}
\end{subfigure}%
%
%
\begin{subfigure}[t]{0.23\textwidth}
\centering
\includegraphics[width=\textwidth]{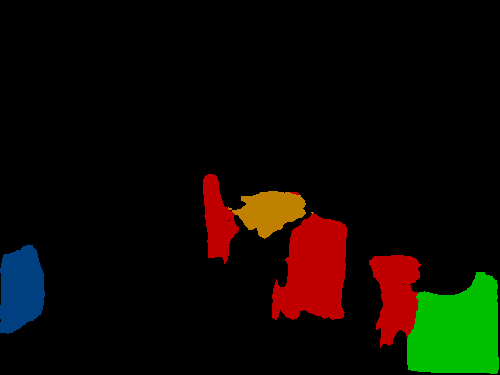}
\label{fig:unattacked_deeplab_prediction}
\end{subfigure}

\begin{subfigure}[t]{0.23\textwidth}
\includegraphics[width=\textwidth]{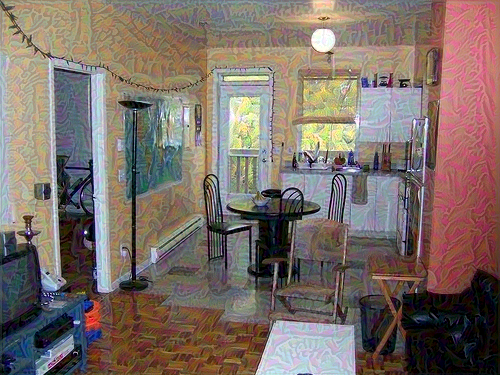}
\label{fig:attacked_deeplab}
\end{subfigure}%
%
%
\begin{subfigure}[t]{0.23\textwidth}
\includegraphics[width=\textwidth]{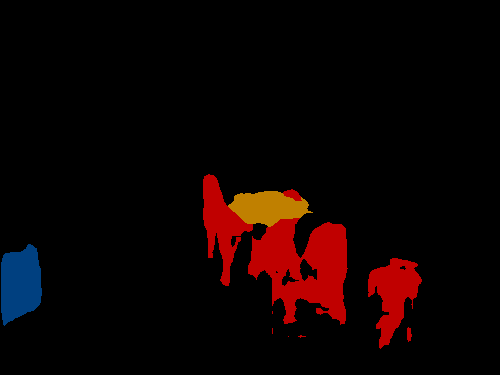}
\label{fig:attacked_deeplab_prediction}
\end{subfigure}

\caption{\textbf{Top row}: output of DeepLabv3+ model~\cite{deeplabv3plus2018} on an image from the Pascal VOC 2012 validation set. \textbf{Bottom row}: output on the same image attacked with A-MI-FGSM. Under an edge-based attack, the segmentation model fails to recognize the sofa.}
\label{fig:deeplabv3_example}
\end{figure}

As shown in Table~\ref{table:deeplab_v3_transfer}, attacks on edge detection also transfer to DeepLabv3+, albeit to a lesser degree. The degradation in this model is smaller than in classification; in Figure~\ref{fig:deeplabv3_example}---a typical example of semantic segmentation---many objects in the scene are still detected. However, like in classification experiments, when we randomly permute the perturbation like~\cite{CXie_2017_ICCV_seg_obj_det}, we observe a much smaller degradation in performance (0.047 drop with permutation and 0.269 without). This demonstrates that the structure of perturbation still matters; the attack cannot be replicated with random noise. 


\begin{table}[t]
\centering
\begin{tabular}{@{}rl@{}}
\toprule
attack       & mIOU  \\ \midrule
none         & 0.822\\
MI-FGSM      & 0.648\\
S-MI-FGSM    & 0.681\\
A-MI-FGSM    & \textbf{0.553}\\
S-\mdifgsm & 0.735\\
A-\mdifgsm & 0.603\\
A-MI-FGSM (permuted) & 0.775\\ \bottomrule
\end{tabular}
\caption{Performance of DeepLabv3+ model on the validation set of PASCAL VOC 2012. The model, \texttt{xception65\_coco\_voc\_train\_aug}, was trained on the COCO and VOC 2012 training datasets (with data augmentation).}
\label{table:deeplab_v3_transfer}
\end{table}

\section{Conclusions}

In this paper, we have added to the wealth of existing evidence that, regardless of task or domain application, undefended deep neural networks are susceptible to adversarial attacks. In particular, we have shown that even a network trained to perform a low-level, ``straightforward" task like edge detection can be confused and manipulated by slight perturbations. This lends further credence to the notion that adversarial examples are intrinsic to current neural networks (or their optimization process) rather than a mere artifact of training data and task. Surprisingly, the same attacks that fool an edge detection network also fool deep networks trained to perform classification and, to a lesser extent, semantic segmentation. These attacks transfer despite significant differences in network architecture and training data.

Still, unresolved questions remain. The exact reasons why edge-based adversarial attacks transfer to classification and segmentation are unclear. Perhaps the low-level cues learned by HED are shared by ImageNet classifiers and semantic segmentation networks, and when these cues are disrupted, all models suffer. Perhaps ImageNet classifiers' reliance on texture information makes them especially prone to certain types of attacks (edge suppression and activation), a question we began to address in Section~\ref{section:stylized_imagenet}.

In this work, we have not explored in depth how to defend against adversarial attacks for edge detection. As a future direction for research, we see potential in protecting HED and similar models against white-box adversarial examples and defending higher-level vision models against edge-based transfer attacks. It is possible that existing defense techniques (\eg, adversarial training~\cite{Goodfellow2014FGSM, madry2018_adversarial_training}) are effective here; otherwise, new defenses may need to be explored.

\section*{Acknowledgements}
We thank Cihang Xie, Yingwei Li, and Wei Shen for their helpful comments on this work. Christian Cosgrove was supported by the 2018 Pistritto Fellowship from the Johns Hopkins Department of Computer Science.
        
{\small

}

\end{document}